\documentclass[sigconf]{acmart}
\usepackage{graphicx}
\usepackage{booktabs}
\usepackage{multirow}
\usepackage{subfig}
\usepackage{hyperref}
\usepackage{pifont}
\usepackage{amsmath}
\usepackage{xcolor}
\usepackage{setspace}
\usepackage[ruled,linesnumbered]{algorithm2e}

\AtBeginDocument{%
  \providecommand\BibTeX{{%
    \normalfont B\kern-0.5em{\scshape i\kern-0.25em b}\kern-0.8em\TeX}}}

\copyrightyear{2022}
\acmYear{2022}
\setcopyright{acmcopyright}
\acmConference[CIKM '22]{Proceedings of the 31st ACM International Conference on Information and Knowledge Management}{October 17--21, 2022}{Atlanta, GA, USA}
\acmBooktitle{Proceedings of the 31st ACM International Conference on Information and Knowledge Management (CIKM '22), October 17--21, 2022, Atlanta, GA, USA}
\acmPrice{15.00}
\acmDOI{10.1145/3511808.3557267}
\acmISBN{978-1-4503-9236-5/22/10}

\begin{document}

\title{Contrastive Representation Learning for Conversational Question Answering over Knowledge Graphs}


\author{Endri Kacupaj}
\email{kacupaj@cs.uni-bonn.de}
\orcid{0000-0001-5012-0420}
\affiliation{%
  \institution{University of Bonn, Germany}
  \country{}
}

\author{Kuldeep Singh}
\email{kuldeep.singh1@cerence.com}
\affiliation{%
  \institution{Zerotha Research and Cerence GmbH, Germany}
  \country{}
}

\author{Maria Maleshkova}
\email{maria.maleshkova@uni-siegen.de}
\affiliation{%
  \institution{University of Siegen, Germany}
  \country{}
}

\author{Jens Lehmann}
\authornote{work done prior to joining Amazon}
\email{lehmann@infai.org} 
\affiliation{%
 \institution{Amazon}
 \institution{InfAI (Institute for Applied Informatics), Germany}
 \country{}
}

\renewcommand{\shortauthors}{Endri Kacupaj, Kuldeep Singh, Maria Maleshkova, \& Jens Lehmann}

\begin{abstract}
   This paper addresses the task of conversational question answering (ConvQA) over knowledge graphs (KGs). The majority of existing ConvQA methods rely on full supervision signals with a strict assumption of the availability of gold logical forms of queries to extract answers from the KG. However, creating such a gold logical form is not viable for each potential question in a real-world scenario. Hence, in the case of missing gold logical forms, the existing information retrieval-based approaches use weak supervision via heuristics or reinforcement learning, formulating ConvQA as a KG path ranking problem. Despite missing gold logical forms, an abundance of conversational contexts, such as entire dialog history with fluent responses and domain information, can be incorporated to effectively reach the correct KG path. This work proposes a contrastive representation learning-based approach to rank KG paths effectively. Our approach solves two key challenges. Firstly, it allows weak supervision-based learning that omits the necessity of gold annotations. Second, it incorporates the conversational context (entire dialog history and domain information) to jointly learn its homogeneous representation with KG paths to improve contrastive representations for effective path ranking. We evaluate our approach on standard datasets for ConvQA, on which it significantly outperforms existing baselines on all domains and overall. Specifically, in some cases, the Mean Reciprocal Rank (MRR) and Hit@5 ranking metrics improve by absolute $10$ and $18$ points, respectively, compared to the state-of-the-art performance.
\end{abstract}



\maketitle

\section{Introduction}
Question answering over knowledge graphs (KGQA) is an essential task that maps a user's utterance to a formal query in order to retrieve the correct answer~\cite{singh2018reinvent}. Recently, with the increasing popularity of intelligent personal assistants (e.g., Alexa, Cortana), the research focus of the scientific community has shifted to Conversational Question Answering over KGs (ConvQA) with multi-turn dialogues~\cite{christmann2019look,kacupaj2021conversational,plepi2021carton}. 

\begin{figure}[!t]
\centering
\captionsetup{type=figure}
\includegraphics[width=0.47\textwidth]{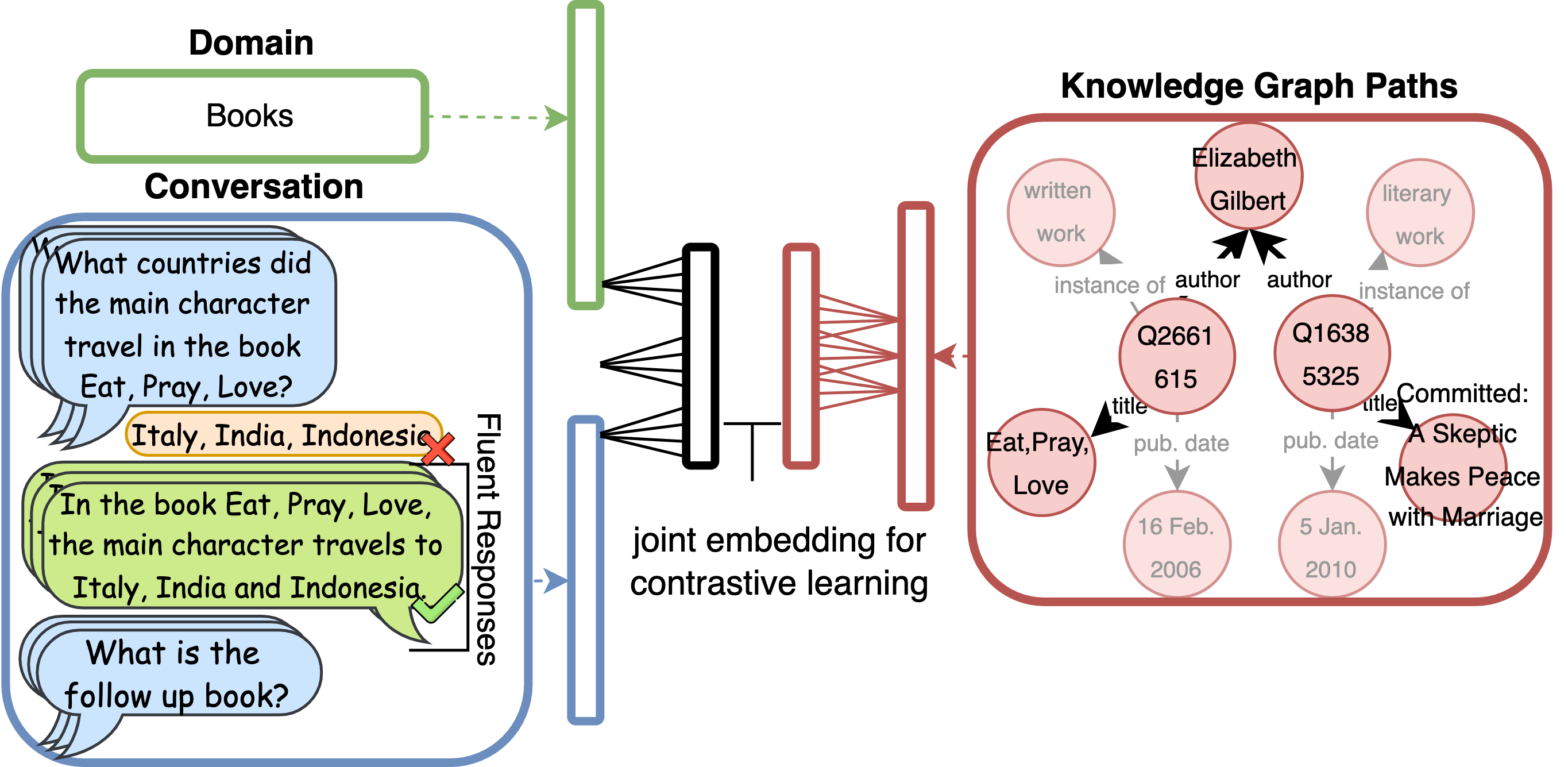}
\caption{Motivating example illustrating a sample conversation~\cite{christmann2019look}. For conversational question answering over KGs, the availability of the entire dialog history with fluent responses and domain information acts as context source in determining the ranking of KG paths while retrieving correct answers. Our proposed approach models conversational context and KG paths in a shared space by jointly learning the embeddings for homogeneous representation.}
\label{fig:intro}
\end{figure}
For the KGQA setup, the existing scientific-literature can be broadly classified into two categories \cite{zaib2021conversational,diefenbach2018core}: (i) semantic parsing approaches, where the goal is to map questions into a logical form, which is then executed over the knowledge graph to extract the correct answers. (ii) Information retrieval approaches aim to retrieve a question-specific graph and apply ranking algorithms to select entities for top positions (i.e., rank KG paths). The two approaches follow either a parse-then execute paradigm or a retrieval-and-rank paradigm. For ConvQA over KGs, there has been significant progress on semantic parsing-based approaches \cite{shen2019multi,kacupaj2021conversational,plepi2021carton}. However, collecting training data for semantic parsing approaches is challenging, and time-consuming \cite{lan2021qasurvey} since each question must be associated with a gold logical form. While for the information-retrieval/ranking-based approaches, only the correct answers (e.g., entities/KG paths) are required for each question. 

\noindent\textbf{State of the Art Limitations.}
Existing ranking-based ConvQA techniques formulate the task as a path ranking problem and propose either a heuristic approach \cite{christmann2019look} or a reinforcement learning model \cite{conquer2021kaiser} to learn from question reformulations (by solely relying on the user). However, these techniques have their inherent limitations. For instance, the rule-based approaches generally require tedious manual labor to generate rules, are error-prone, and are prone to bias \cite{chiticariu2013rule}. Furthermore, existing scientific literature points to several limitations in solely relying on users for query reformulation, and focused on automatic query suggestion/related query recommendations techniques \cite{chen2021towards,chen2021incorporating}. Firstly, entirely relying on users creates a heavy cognitive load on them \cite{na2021effects}. Secondly, reformulation query issued within a time interval that is short enough (such as five minutes) implies dissatisfaction \cite{hassan2013beyond}. Lastly, a recent study points out that critical issues for conversational systems
concerning reformulations have not been widely analyzed in the literature, especially the reformulations’ implicit
ambiguity \cite{kiesel2021toward}.
Hence, for advancing IR-based ConvQA methods, there is a desirable ask for approaches that are independent of availability of gold-logical form, heuristic rules, or its sole dependency on the users for query reformulation. 
\subsection{Proposed Approach, Motivation, and Contributions}
This paper addresses a relatively unexplored research area: the ranking-based ConvQA task for answering conversational questions posed against a knowledge graph. In this context, we propose PRALINE (\textbf{P}ath \textbf{R}anking for convers\textbf{A}tiona\textbf{L} quest\textbf{I}on a\textbf{N}sw\textbf{E}ring), a novel contrastive representation learning approach to rank KG paths for retrieving the correct answers effectively. Furthermore, we enrich the learning process by incorporating the available conversational context, i.e., ~(1) the entire dialog history with (2) fluent responses and (3) domain information (c.f. Figure \ref{fig:intro}).
Our rationale for critical choices are following: 

\noindent\textbf{Why Contrastive Learning for ConvQA?}\\
Contrastive learning \citep{chopra2005learning,radford2021learning} aims at learning representations of data by contrasting similar and dissimilar samples.
For our task, the data contains conversations and respective answers (label) for each conversational question. However, there is no discrete information on how the answers were extracted. Therefore fully supervised learning approaches such as semantic parsing \cite{shen2019multi,plepi2021carton} cannot be directly applied without annotations. Consequently, we can design an approach with contrastive learning by only augmenting the data with KG paths that lead to the correct answers. These KG paths have as starting points the context entities mentioned in the conversation and landing points the answers. Furthermore, extracted KG paths leading to correct answers are marked as positive, while others are negative. In this way, contrastive learning is ideal for our task since it allows us to rank KG paths that are considered positive and answer conversational questions.

\noindent\textbf{Why Conversational Context to Enrich Learning Process?}~\\
Conversational context plays a vital role in human understanding~\cite{doyle2007role} and question answering~\cite{lin2003makes}. To further enhance learning, we seek additional conversational context to improve the ranking performance and allow contrastive learning approach to distinguish between positive and negative paths. Such information can be the conversation domain and fluent natural language answers \cite{kacupaj2020vquanda,biswas2021vanilla,kacupaj2021paraqa,kacupaj2021vogue} instead of standalone KG answers (answer labels). 
Identifying the domain of the conversation allows us to enrich the representations and efficiently contrast negative paths that do not use properties of the particular domain. Moreover, fluent natural language answers will supplement the conversations with additional textual context to support the learning and ranking process. 

\noindent\textbf{Contributions}:
We make the following key contributions in the paper:
1) We propose PRALINE, the first contrastive learning based approach for ConvQA that jointly models the available conversational context (full dialog history with fluent responses and domain) and KG paths in a common space for learning joint embedding representations to improve KG path ranking. 
2) We systematically study the impact of incorporating additional context on the performance of PRALINE. Results on standard datasets show a considerable improvement over previous baselines. 
To facilitate reproducibility and reuse, our framework implementation and the results are publicly available\footnote{\url{https://github.com/endrikacupaj/PRALINE}}.
The structure of the paper is as follows: Section~\ref{sec:related_work} summarizes the related work. Section~\ref{sec:task_definition} provides the concepts, notations and tasks definitions. Section~\ref{sec:approach} presents the proposed PRALINE framework. Section~\ref{sec:experiment} describes the experiments, including the experimental setup, the results, the ablation study and error analysis. We conclude in Section~\ref{sec:conclusion}.
\section{Related Work}\label{sec:related_work}
Considering KGQA is a widely studied research topic, we stick to the work closely related to our proposed approach (detailed surveys are in \cite{zaib2021conversational,diefenbach2018core}).

\noindent\textbf{Single-shot KGQA.} 
Several KGQA works handle the task as a semantic graph generation and re-ranking. \citet{bast2015kgqa} compare a set of manually defined query templates against the natural language question and generate a set of query graph candidates by enriching the templates with potential relations. \citet{yih2015semantic} creates grounded query graph candidates using a staged heuristic search algorithm and employs a neural ranking model to score and find the optimal semantic graph. \citet{yu2017improved} use a hierarchical representation of KG relations in a neural query graph ranking model. Authors compare the results against a local sub-sequence alignment model with cross attention~\cite{parikh2016decomposable}. \citet{maheshwari2019learning} conduct an empirical investigation of neural query graph ranking approaches by experimenting with six different ranking models. The proposed approach is a self-attention-based slot matching model that exploits the inherent structure of query graphs.

\noindent\textbf{ConvQA over KGs.}
Most recent works on ConvQA \citep{shen2019multi,kacupaj2021conversational,plepi2021carton} employ the semantic parsing approach to answer conversational questions. The first work in this area \citet{saha2018complex} propose a hybrid model of the HRED model \cite{serban2016dialogue} and the key-value memory network model \cite{miller2016key}. The model consists of three components; where the first one is the hierarchical encoder, which computes the utterance representation. The next module is a higher-level encoder that computes the context representation. The second component is the Key-Value Memory Network that stores each candidate tuples as a key-value pair. The key contains the concatenated embedding of the relation and the subject. The last component is the decoder used to create an end-to-end solution and produce multiple types of answers. Other approaches extend similar idea, although using multi-task learning paradigm \cite{kacupaj2021conversational,plepi2021carton}.
\citet{christmann2019look} proposes an approach that answers conversational questions over a knowledge graph (KG) by maintaining conversation context using entities and predicates seen so far and automatically inferring missing or ambiguous pieces for follow-up questions. The core of this method is a graph exploration algorithm that judiciously expands a frontier to find and rank candidate answers for the given question.
\citet{conquer2021kaiser} present a reinforcement learning model that can learn from a conversational stream of questions and reformulations. Authors model the answering process as multiple agents walking in parallel on the KG, where the walks are determined by actions sampled using a policy network. The policy network takes the question and the conversational context as inputs and is trained via noisy rewards obtained from the reformulation likelihood. Our work lies closely with \citet{christmann2019look,conquer2021kaiser}. However, these approaches focus either on the rule-based method or require explicit feedback from (non-expert) users creating additional issues/dependencies such as higher cognitive load on the user, potential bias, and error-prone. Further, these approaches ignore the entire dialog history and do not consider fluent responses as contextual sources. Therefore, our focus is to explore a weak supervision method that relies solely on conversation context and available KG candidate paths to retrieve the final answer.

\noindent\textbf{Contrastive Learning Approaches.}
Early work \cite{chopra2005learning} in contrastive learning introduces the concept of contrastive loss for facial recognition. It takes a pair of inputs and aims for minimal embedding distance when they are from the same class but maximizes the distance otherwise. Later, there were several approaches extended contrastive learning to different use-cases such as identifying image captions \cite{radford2019better}, computing code from an augmented image \cite{caron2020unsupervised}, feature clustering \cite{caron2018deep}, and dense information retrieval \cite{izacard2021towards}. In our setting, the challenge is to adapt contrastive learning to 
compute a joint loss between conversation utterances, their additional context (fluent responses, domain), and candidate KG paths. Now we detail how to address these challenges in our proposed approach. 
\section{Concepts, Notation and Problem Formulation}\label{sec:task_definition}
We define a KG as a tuple $\mathcal{K} = (\mathcal{E}, \mathcal{R}, \mathcal{T}^{+})$ where $\mathcal{E}$ denotes the set of entities (vertices), $\mathcal{R}$ is the set of relations (edges), and $\mathcal{T}^{+} \subseteq \mathcal{E} \times \mathcal{R} \times \mathcal{E}$ is a set of all triples. A triple $\tau = (e_h, r_{h,t}, e_t) \in \mathcal{T}^{+}$ indicates that, for the relation $r_{h,t} \in \mathcal{R}$, $e_h$ is the head entity (origin of the relation) while $e_t$ is the tail entity. For a KG, a conversation $\mathcal{C}$ with $\mathrm{T}$ turns is composed from a set of a sequence of questions $\mathcal{Q} = \{ q^t \}$ and corresponding answers $\mathcal{A} = \{ a^t \}$, where $t = 0, 1,...,\mathrm{T}$, such that $\mathcal{C} = \langle (q^0, a^0), (q^1, a^1),\allowbreak ..., (q^\mathrm{T}, a^\mathrm{T}) \rangle$. Furthermore, each question $q^t$ is a sequence of tokens $q^{t}_{i}$, such that $\langle q^{t}_{1}, ..., q^{t}_{|q^{t}|} \rangle$, where $|q^{t}|$ is the number of tokens in $q^{t}$. For each question $q^{t}$ we have a conversation history $\mathcal{C}^t$, where for question $q^0$ the conversation history is $\emptyset$. We define answer $a^t$ for question $q^t$ is a set of entities or literals in $\mathcal{K}$. 
We define fluent responses as $v^t$, which is a sequence of tokens $v^{t}_{i}$, where $a^t \in v^t$. Similar to question $q^t$ we have $\langle v^{t}_{1}, ..., v^{t}_{|v^{t}|} \rangle$, where $|v^{t}|$ is the number of tokens in $v^{t}$. With fluent answers the conversation $\mathcal{C}$ can be illustrated as $\mathcal{C} = \langle (q^0, v^0), (q^1, v^1), ..., (q^\mathrm{T}, v^\mathrm{T}) \rangle$.

We define the following two concepts explicitly:

\noindent\textbf{Context Entities.}
The set of context entities $\mathcal{E}_{c} \subseteq \mathcal{E}$ contains entities mentioned in question $q^t$, answer $a^t$ and conversation $\mathcal{C}$. 

\noindent\textbf{Context Paths.}
Context KG paths $\mathcal{P}_{c}$ are extracted given the context entities $\mathcal{E}_{c}$, where $\mathcal{P}_{c} \subseteq \{\mathcal{E}_{c} \times \mathcal{R} \times \mathcal{E}\} + \{\mathcal{E}_{c} \times \mathcal{R} \times \mathcal{E} \times \mathcal{R} \times \mathcal{E}\} + \{\mathcal{E}_{c} \times \mathcal{R} \times \mathcal{E} \times \mathcal{R} \times \mathcal{E} \times \mathcal{R} \times \mathcal{E}\}$. This means that extracted KG paths $\mathcal{P}_{c}$ will be either 1-hop, 2-hop or 3-hop paths where all of them start with context entities $\mathcal{E}_{c}$. For a question $q^t$, we define $\mathcal{D}^{t+} \subseteq \mathcal{P}^{t}_{c}$ and $\mathcal{D}^{t-} \subseteq \mathcal{P}^{t}_{c}$ the set of positive/correct and negative/incorrect context paths, respectively. Where, $\mathcal{D}^{t+} \cup \mathcal{D}^{t-} = \mathcal{P}^{t}_{c}$.

\noindent\textbf{Problem Formulation.}
For the ConvQA task, given a knowledge graph $\mathcal{K}$, a natural language question $q^t$, the conversation history $\mathcal{C}^{t}$, and the set of context entities $\mathcal{E}_{c}^{t}$, the goal is to extract all the potential KG paths $\mathcal{P}_{c}^{t}$. We formulate the ConvQA task as an IR problem where we score and rank $\mathcal{P}_{c}^{t}$ to select top context paths $p_{c}^{t} \in \mathcal{P}_{c}^{t}$ which leads us to entities or literals that match the gold answer $a^t$, which is also the answer of the question $q^t$.
We achieve this by employing contrastive representation learning, which aims to learn an embedding space where similar sample pairs have representations close to each other while dissimilar ones are far apart.

\begin{table}[!t]
\begin{tabular}{ll}
\toprule
\textbf{Notation} & \textbf{Concept} \\
\midrule
$\mathcal{K}, \mathcal{E}, \mathcal{R}, \mathcal{T}^{+}$ & Knowledge Graph, entities, relations, triples \\
$\mathcal{C}, t$& Conversation, turn \\
$q^t, a^t$ & Question and answer at turn t \\
$v^t$ & Fluent response at turn t \\
$\tau^t$ & Domain information at turn t \\ 
$\mathcal{C}^t$ & Conversation history at turn t \\
$\mathcal{E}_{c}, \mathcal{P}_{c}$ & Context entities, context KG paths \\
$\mathcal{D}^{t+}, \mathcal{D}^{t-}$ & Set of positive and negative context paths for $q^t$ \\ \midrule
$s^t$ & Input sequence (contains $\mathcal{C}^t$ and $q^t$) \\ 
$d$ & Space dimension \\ 
$h^{(\cdot)}$ & Contextual embeddings \\ 
$\theta^{(\cdot)}$ & Trainable parameters \\ 
$\boldsymbol{W}^{(\cdot)}$ & Weight matrix for linear layer \\ 
$\omega^{(\cdot)}$ & Probability distribution over vocabulary \\ 
$\phi^{c}, \phi^{p}$ & Joint embeddings for conversation and path \\ 
\bottomrule
\end{tabular}
\caption{Notation for concepts in PRALINE.}
\label{tab:notations}
\end{table}

\begin{figure*}[!t]
\centering
\captionsetup{type=figure}
\includegraphics[width=0.75\textwidth]{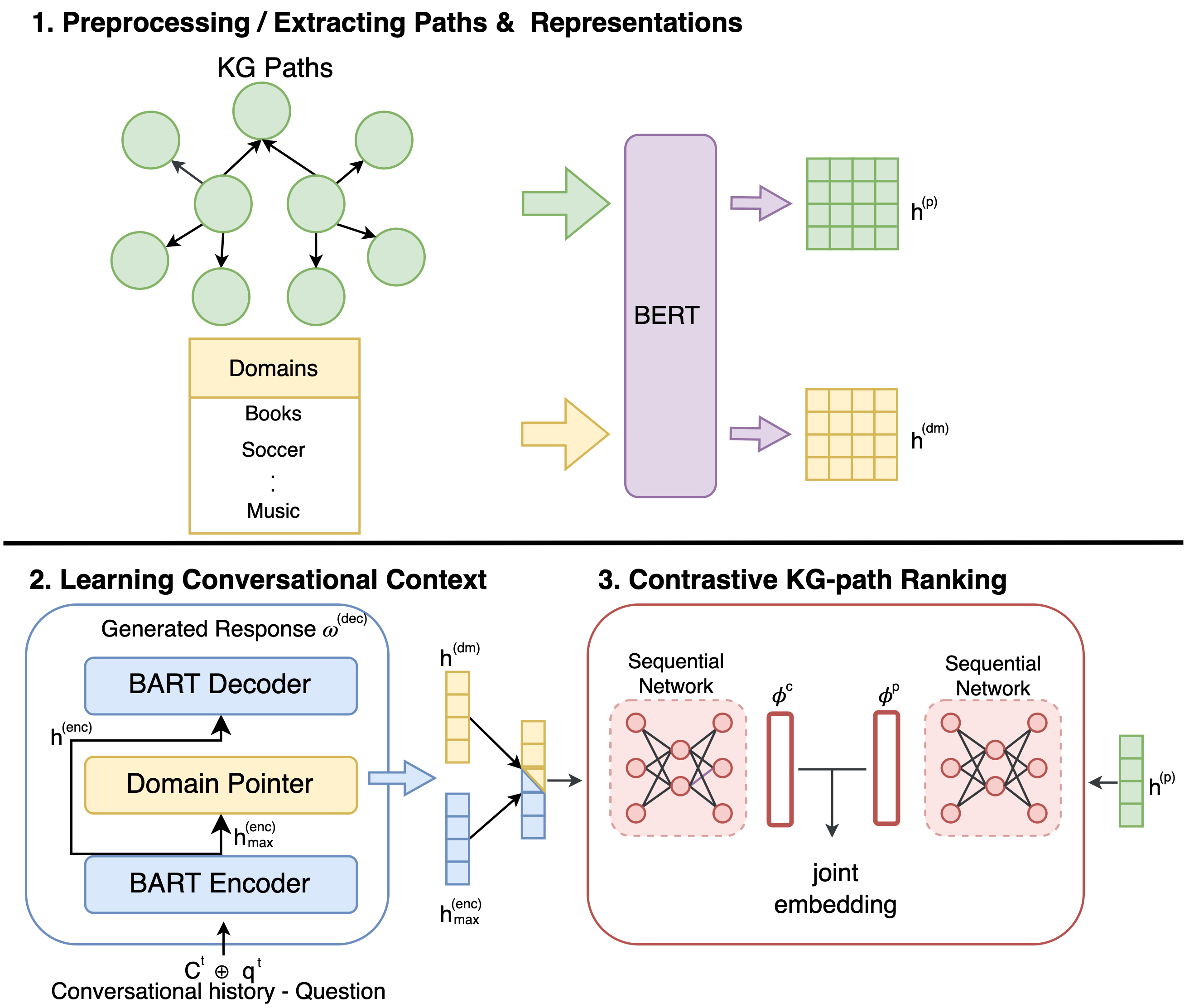}
\caption{PRALINE (\textit{P}ath \textit{R}anking for convers\textit{A}tiona\textit{L} quest\textit{I}on a\textit{N}sw\textit{E}ring) architecture. It consists of three steps: 1) Extract KG paths and domains and represent them using a BERT model. 2) Learn the conversational context using a BART model and a domain identification pointer. 3) A contrastive ranking module that learns a joint embedding space $\phi^{c}, \phi^{p}$ for the conversation (contextual embeddings $h^{(enc)}$ \& selected domain embeddings $h^{(dm)}$) and the context path $h^{(p)}$.}
\label{fig:architecture}
\end{figure*}

\section{Approach}\label{sec:approach}
In a conversation, the input data consists of questions $q^t$ and answers $a^t$, extracted from the knowledge graph. We propose a contrastive learning approach PRALINE to rank KG paths. In particular, PRALINE ranks (KG) context paths $\mathcal{P}_{c}^{t}$ depending on the particular utterance in a conversation. 
Figure~\ref{fig:architecture} illustrates how PRALINE operates. 
It consists of three step processes: 1) first step preprocess and extract potential candidate paths (and their domain). 2) the second step collects and encodes possible conversation context. The available answer labels do not provide much information to enrich the learning process; therefore, fluent responses are utilized. 3) Last step jointly embeds conversation, its context, and candidate KG paths in a common space to apply a contrastive ranking module to rank KG paths effectively. 

The last two steps are jointly trained in an end-to-end manner, post preprocessing.
Furthermore, our approach does not explicitly depends on conversation context, and this context is used as additional signals. As we observe in experiments, such contexts can be omitted (cf. Table \ref{tab:ablation}). Also, the implementation choices to encode either the embeddings during preprocessing steps (e.g., BERT \cite{devlin2019bert}) or domain identifier (for example, pointer network \cite{vinyals2015pointer} ) are used for empirical effectiveness using state-of-the-art techniques proposed to solve individual sub-tasks in the ConvQA task. Hence, the approach can be implemented using other implementation choices, as we see in ablation later in the section \ref{sec:experiment}. We now detail the approach. 

\subsection{Preprocessing \& Extracting Representations}
For our approach, we identify context entities $\mathcal{E}_c$ and extract potential candidates for KG paths similar to~\citet{conquer2021kaiser} and do not claim as our novelty. 
After extracting the KG paths $\mathcal{P}_{c}$, we initialize their representations using sentence embeddings that implicitly employ underlying hidden states from BERT network~\cite{devlin2019bert}. We treat each KG path as a sentence and feed that as an input to BERT. 
The KG path representations are used for the contrastive ranking task. 
Similarly, we preprocess and embed the domains of the conversations to generate the representation $h^{(dm)}$. The embedded domains are implicitly utilized during the ranking process. 

\textbf{Please note} that this step is not part of the trainable architecture we describe in the next two steps, and models such as BERT are only used during preprocessing phase.
\subsection{Learning Conversational Context}

\textbf{Encoding Conversation History and Question}~\\
As the first step of our framework, we utilize a BART-based bidirectional encoder~\citep{lewis2020bart} in order to encode both the conversation history $\mathcal{C}^t$ and current question $q^t$ at turn $t$. The conversation history also contains fluent responses from previous turns (e.g. $v^{t-1}, v^{t-2},...$). Here we concatenate the conversation history $\mathcal{C}^t$ and current question $q^t$ using a helper token $[SEP]$ to create the input sequence $s^t = \mathcal{C}^t \; \oplus [SEP] \oplus \; q^t$, where $\oplus$ is the operation of sequence concatenation.\footnote{With the same $[SEP]$ token, the questions and fluent answers are separated inside the conversation history.}
Next, we tokenize the input sequence $s^t$ into $|s^t|$ tokens $\{s^{t}_1,...,s^{t}_{|s^t|}\}$, where $|s^t| = |\mathcal{C}^t| + |q^t| + 1$, using a byte-level Byte-Pair-Encoding tokenizer.
Then, we forward the tokenized sequence into the encoder and it outputs the contextual embeddings $h^{(enc)} =  \{h_{1}^{(enc)},\dots,h_{n}^{(enc)}\}$, where $h_{i}^{(enc)} \in \mathbb{R}^{d}$, $d$ is the space dimension, $i \in \{1,...,n\}$ and $n=|s^t|$. We define the encoder as: 
\begin{equation}
\begin{split}
    &h^{(enc)} = encoder(x; \theta^{(enc)}),
\end{split}
\end{equation}
\noindent where $\theta^{(enc)}$ are the encoder's trainable parameters.

Considering that fluent responses provide additional textual context compared to answer labels, we enable PRALINE to generate fluent responses and employ them for forthcoming questions. This justifies why PRALINE utilizes BART, a sequence-to-sequence model pretrained to reconstruct/generate text. 
For generating the response $v^t$ we employ a BART-based decoder where we provide the encoder contextual embeddings $h^{(enc)}$ of the input sequence $s^t$. The decoder vocabulary is defined as: 
\begin{equation}
    V^{(dec)} = V^{(v)} \; \cup \; \{\; [ANS] \; \},
\end{equation}
\noindent where $V^{(v)}$ is the vocabulary with all the distinct tokens from the generation task. We also employ an additional helper token to specify the position of the answer $a^t$ on the final generated sequence.
A linear layer and a softmax follow the decoder in order to calculate each token's probability score in the vocabulary. We define the decoder stack output as follows:
\begin{equation}
\begin{split}
    &h^{(dec)} = decoder(h^{(enc)};\theta^{(dec)}),\\
    &\omega_{i}^{(dec)} = softmax(\boldsymbol{W}^{(dec)} h_{i}^{(dec)}),
\end{split}
\end{equation}
\noindent where $h_{i}^{(dec)}$ is the hidden state in time step $i$, $\theta^{(dec)}$ are the decoder trainable parameters, $\boldsymbol{W}^{(dec)} \in \mathbb{R}^{|V^{(dec)}|\times d}$ are the linear layer weights, and $\omega_{i}^{(dec)} \in \mathbb{R}^{|V^{(dec)}|}$ is the probability distribution over the decoder vocabulary in time step $i$. The $|V^{(dec)}|$ denotes the decoder's vocabulary size.\\

\noindent\textbf{Domain Identification Pointer}\label{sec:domain}~\\
The second step is a domain identification pointer network. This module is responsible for identifying the KG domain of the input sequence $s^t$ and employs a pointer architecture inspired from \citet{vinyals2015pointer}. In general, pointer networks are robust to handle different vocabulary sizes for each time step \cite{vinyals2015pointer} which was our rationale for their integration in PRALINE. 
Another advantage of using pointer networks compared to simple classifiers is that the vocabulary of the domains can be updated during evaluation or inference.

We define the vocabulary as $V^{(dm)} = \{\tau_1,\dots,\tau_{n_{dm}}\}$, where $n_{dm}$ is the total number of domains in the KG. To compute the pointer scores for each domain candidate, we use the encoder contextual embeddings $h^{(enc)}$. We model the pointer networks with a feed-forward linear network followed by a softmax layer. 
We can define the domain pointer as: 
\begin{equation} \label{pointer-omega}
     \omega_{i}^{(dm)} = softmax(\boldsymbol{W}_1^{(dm)}u_t^{(dm)}),
\end{equation}
\noindent where $\omega_{i}^{(dm)} \in \mathbb{R}^{|V^{(dm)}|}$ is the probability distribution over the domain vocabulary. The weight matrix $\boldsymbol{W}_1^{(dm)} \in \mathbb{R}^{1 \times d_{kg}}$. Also, $u_t^{(dm)}$ is a joint representation that includes the domain embeddings and the contextual embeddings, computed as:
\begin{equation}
    u_{t}^{(dm)} = tanh(\boldsymbol{W}_2^{(dm)}\tau + h^{(enc)}),
\end{equation}
\noindent where the weight matrix $\boldsymbol{W}_2^{(dm)} \in \mathbb{R}^{d \times d_{kg}}$. We denote with $d_{kg}$ the dimension used for domain (KG) embeddings.\footnote{For our experiments we employ $d_{kg}$ = $d$} $\tau \in \mathbb{R}^{d_{kg} \times |V^{(dm)}|}$ are the domain embeddings. $tanh$ is the non-linear layer.

\subsection{Contrastive KG-path Ranking}\label{sec:ranking}
We propose a contrastive ranking module by employing two identical sequential networks in order to generate joint embeddings for a conversation (input sequence $s^t$) and a context path $p_{c}^{t}$ at turn $t$. 
Each sequential network contains two linear layers separated with a $ReLU$ activation function, and appended with a $tahn$ non-linear layer. 
Here, as input we consider the concatenation of the encoder contextual embeddings $h^{(enc)}$ together with the embedded domain selected from the domain identification pointer (cf. equation \ref{pointer-omega}). In this way we incorporate also the domain information when we create the joint embeddings. For a conversation, the encoder contextual embeddings are $h^{(enc)}$ where $h^{(enc)} \in \mathbb{R}^{|s^t| \times d}$. The contextual embeddings contain representations of dimension space $d$ for each token of the input sequence $s^t$. While, for the domain embedding $h^{(dm)}$ and for each context path embedding $h^{(p)}$ (both initialized using BERT embeddings), we have $h^{(dm)} \in \mathbb{R}^{d_{kg}}$ and $h^{(p)} \in \mathbb{R}^{d_{kg}}$, respectively (implementation details explained in \ref{sec:setup}). In order to match the space dimensions $\mathbb{R}^{|s^t| \times d}$ and $\mathbb{R}^{d}$, a $max$ layer is applied to the encoder contextual embeddings $h^{(enc)}$ before forwarding it to the sequential network of the module.
We define this as:
\begin{equation}
    h^{(enc)}_{max} = max_{0} h^{(enc)},
\end{equation}
\noindent where $max_{0}$ indicates the max operation performed in dimension $zero$, and $h^{(enc)}_{max} \in \mathbb{R}^{d}$.
Overall, we define the module sequential networks as:
\begin{equation}
\begin{split}
    \phi^{c} &= tahn(\boldsymbol{W}^{(crk)}_2 ReLU(\boldsymbol{W}^{(crk)}_1 [h^{(enc)}_{max}; h^{(dm)}])), \\
    \phi^{p} &= tahn(\boldsymbol{W}^{(prk)}_2 ReLU(\boldsymbol{W}^{(prk)}_1 h^{(p)})),
\end{split}
\end{equation}
\noindent where $\boldsymbol{W}^{(crk)}_1 \in \mathbb{R}^{d\times 2d}$, $\boldsymbol{W}^{(prk)}_1 \in \mathbb{R}^{d\times d}$ are the weight matrices for the first linear layers. $\boldsymbol{W}^{(crk)}_2 \in \mathbb{R}^{d\times d}$, $\boldsymbol{W}^{(prk)}_2 \in \mathbb{R}^{d\times d}$ are the weight matrices for the second linear layers. $\phi^{c} \in \mathbb{R}^{d}$ and $\phi^{p} \in \mathbb{R}^{d}$ are the final joint embeddings on space dimension $d$ of the conversation and context path, respectively.

Given a batch of (conversational context, KG paths) pairs during training, this module computes the cosine similarity between all possible candidates within this batch. The conversation and KG path sequential networks are jointly trained to maximize the similarity between the correct pairs while minimizing the similarity for incorrect pairs.


\begin{algorithm}[!t]
\KwIn{Training set $S_{train} = \{ (q^{t}, C^{t}, v^t, \tau^t, \mathcal{D}^{t+}_{c}, \mathcal{D}^{t-}_{c}) \}$}
\BlankLine
\For{$S_{batch} \in S_{train}$}{ 
    $q_{b} \leftarrow getQuestions(S_{batch})$\\
    $C_{b} \leftarrow getConvHistory(S_{batch})$\\
    $v_{b} \leftarrow getFluentAnswers(S_{batch})$\\
    $\tau_{b} \leftarrow getDomains(S_{batch})$\\
    
    \BlankLine
    $y^{(rk)}_{b}$ : $y^{(rk)}_{b} \in \mathbb{R}^{\{1, -1\} \times b}$\\
    \BlankLine
    
    \For{$y^{(rk)}_{i} \in y^{(rk)}_{b}$}{ 
        \uIf{$y^{(rk)}_{i} = 1$}{
        $\mathcal{D}^{+} \leftarrow getPositivePaths(S_{batch})$\\
        $p_{i} \sim samplePath(\mathcal{D}^{+})$
        }
        
        \Else{
            $\mathcal{D}^{-} \leftarrow getNegativePaths(S_{batch})$\\
            $p_{i} \sim samplePath(\mathcal{D}^{-})$
        }
    }
    
    \BlankLine
    $h^{(p)}_{b} \leftarrow embedPaths(p_b)$\\
    $h^{(dm)}_{b} \leftarrow embedDomains(\tau_b)$\\
    \BlankLine
    
    \Begin(PRALINE forward){
        $h^{(enc)}_{b} \leftarrow PRALINE.encoder(q_b, C_b)$\\
        $\omega^{(dec)}_{b} \leftarrow PRALINE.decoder(h^{(enc)}_{b}, v_{b})$
        \BlankLine
        
        $\omega^{(dm)}_{b} \leftarrow PRALINE.domainPointer(h^{(enc)}_{b})$\\
        \BlankLine
        
        $\phi^{c}_{b}, \phi^{p}_{b} \leftarrow PRALINE.ranking(h^{(enc)}_{b}, h^{(dm)}_{b}, h^{(p)}_{b})$\\
    }
    
    $L^{dm}_{b} = \frac{1}{b} \sum_{i=1}^{b} - \sum_{j=1}^{m} log p(y_{j}^{(dm)} | \omega_{i}^{(dm)})$\\
    \BlankLine
    $L^{dec}_{b} = \frac{1}{b} \sum_{i=1}^{b} - \sum_{k=1}^{n} log p(y_{l}^{(dec)} | \omega_{i}^{(dec)})$\\
    \BlankLine
    \BlankLine
    $L^{rk}_{b} = \frac{1}{b} \sum_{i=1}^{b} \begin{cases}
                1 - cos(\phi^{c}_{i}, \phi^{p}_{i}), & \text{if } y^{(rk)}_{i} = 1\\
                max(0, cos(\phi^{c}_{i}, \phi^{p}_{i}) - \alpha),              & \text{if } y^{(rk)}_{i} = -1
              \end{cases}$\\
    \BlankLine
    
    Update PRALINE weights w.r.t. $\;\lambda_1 L^{dm}_{b} + \lambda_2 L^{rk}_{b} + \lambda_3 L^{dec}_{b}$
}
\caption{Proposed Algorithm of PRALINE}
\label{alg:learning}
\end{algorithm}

\subsection{Joint Contrastive Learning}
PRALINE consists of three trainable modules for which a loss function is applied. The encoder is trained based on the signal received from the domain identification pointer, contrastive ranking module, and decoder. For training simultaneously all the modules/tasks, we perform a weighted average of all the single losses as follows:
\begin{equation}
    L = \lambda_1 L^{dm} + \lambda_2 L^{rk} +  \lambda_3 L^{dec},
\end{equation}
\noindent where $\lambda_1, \lambda_2, \lambda_3$ are the relative weights. $L^{dm}$ and $L^{dec}$ are the respective negative log-likelihood losses of the domain identification pointer and decoder modules. While the $L^{rk}$ is the cosine embedding loss for the ranking module. 
These losses are defined as: 
\begin{equation}
\begin{split}
    &L^{dm} = - \sum_{j=1}^{m} log p(y_{j}^{(dm)} | x), \\
    &L^{rk} = \begin{cases}
                1 - cos(\phi^{c}, \phi^{p}), & \text{if } y^{(rk)} = 1\\
                max(0, cos(\phi^{c}, \phi^{p}) - \alpha),              & \text{if } y^{(rk)} = -1
              \end{cases}, \\
    &L^{dec} = - \sum_{l=1}^{n} log p(y_{l}^{(dec)} | x),
\end{split}
\end{equation}
\noindent where $n$ is the length of the gold fluent response. 
$y_{j}^{(dm)} \in V^{(dm)}$ are the gold labels for the domain identification pointer and
$y_{l}^{(dec)} \in V^{(dec)}$ are the gold labels for the decoder. $y^{(rk)} \in \{ 1, -1 \}$ are the gold labels for the ranking module. $cos(\cdot)$ is the normalized cosine similarity and $\alpha$ is the margin.
The model benefits from each module's supervision signals, which improves the performance in the given task (cf. section \ref{sec:experiment}). Algorithm~\ref{alg:learning} illustrates high-level pseudo-code for PRALINE's learning process.

\subsection{Inference}
Once training is complete, for inference, we first identify the context entities $\mathcal{E}^c$ in the input question $q^t$ and conversational history $\mathcal{C}^t$. 
Using the context entities $\mathcal{E}^c$, we extract the context paths $\mathcal{P}^c$ from the knowledge graph. 
After we follow the following four steps: i) encode the input conversation and question, ii) identify the domain of the encoded conversation, iii) employ both domain information and encoded conversation to score all candidate paths via their cosine similarity and rank them to retrieve the answer from the highest scored path. iv) Using the encoded conversation and the retrieved response, generate the fluent response via the decoder.

\section{Experiments}\label{sec:experiment}
\subsection{Experimental Setup} \label{sec:setup}
\textbf{Model Configurations.}
Table~\ref{tab:hyper} summarizes the hyperparameters used for the experiments. For all the modules in the PRALINE framework, we employ a space dimension $d=768$. We utilize a BART (base) \cite{lewis2020bart} model for the encoder and decoder.
For training parameters, we employ a batch size of $32$, a learning rate of $1e-4$, and we train for $120$ epochs and store the models' checkpoints. For the optimization, we use the AdamW algorithm with weight decay fix as introduced in \cite{loshchilov2018decoupled}. 
We apply residual dropout in different parts and modules of our framework (such as domain pointer and ranking) with a probability of $0.1$. 
As mentioned above, we use BERT pre-trained model for generating the initial embeddings of the domains and KG paths. 
We also restrict PRALINE's input sequence ($C^t$ + $q^t$) size to $150$ tokens. For the domain pointer and decoder negative log-likelihood losses, we apply relative weights $\lambda_1$ and $\lambda_2$ of $0.25$. Finally, for the path ranking cosine embedding loss, we use a margin $\alpha$ of $0.1$ and relative weight $\lambda_3$ of $1.0$.

\begin{table}[!t]
\small
\centering
\begin{tabular}{lc}
\toprule
\textbf{Hyperparameters} & \textbf{Value} \\ \midrule
epochs & $120$ \\
batch size & $32$ \\
learning rate & $1e-4$ \\
dropout ratio & 0.1 \\
optimizer & AdamW \\
model dim & $768$ \\
$v^t$ max length & $50$ \\
$C^t$ + $q^t$ max length & $150$ \\
domain pre-trained embeddings & BERT \\
KG paths pre-trained embeddings & BERT \\
$\lambda_1$, $\lambda_2$, $\lambda_3$ & $0.25$, $1.0$, $0.25$ \\
margin $\alpha$ & $0.1$ \\
\bottomrule
\end{tabular}
\caption{Hyperparameters for PRALINE.}
\label{tab:hyper}
\vspace{-1.5em}
\end{table}

\noindent\textbf{Datasets and Models for Comparison.}
For ConvQA over KGs, we compare our framework on two relevant datasets ConvQuestions~\cite{christmann2019look} and ConvRef~\cite{conquer2021kaiser}. We further employed fluent responses for both datasets~\cite{kacupaj2022answer}.
Our first baseline is CONVEX~\cite{christmann2019look} which detects answers to conversational utterances over KGs in a two-stage process based on judicious graph expansion. First, it detects frontier nodes that define the context at a given turn. Then, it finds high-scoring candidate answers in the vicinity of the frontier nodes. 
The second baseline and current state-of-the-art is CONQUER~\cite{conquer2021kaiser}, an RL-based method for conversational QA over KGs, which leverages implicit negative feedback when users reformulate previously failed questions. A recently proposed model OAT \cite{marion2021structured} reports values on ConvQuestions that proposes a semantic parsing-based approach. Focal entity \cite{lan2021modeling} is another recently release baseline. For ConvQuestions, we took baseline values from official leaderboard and for ConvRef, values are from the baseline papers. 

\noindent\textbf{Evaluation Metrics.}
For evaluating the ConvQA performance, we use the following ranking metrics which are also employed by the previous baselines: i)
Precision at the top rank (P@1) 2) Mean Reciprocal Rank (MRR) is the average across the reciprocal of the rank at which the first context path was retrieved. 3) Hit at 5 (H@5) is the fraction of times a correct answer was retrieved within the top-5 positions.
We report precision, recall, and F1-score for the domain identification task, while for response generation, we employ BLEU-4 and METEOR.

\subsection{Results}
\textbf{Research Questions:} We conduct our experiments and analysis in response to the question \textbf{RQ:} For a given conversation utterance, what is the efficacy of contrastive learning approach implemented in PRALINE for ranking the KG paths? As such, our research question is further divided into sub-research questions \textbf{RQ1.1:} what is the effect of conversational context on the efficiency of PRALINE? \textbf{RQ1.2:} what is the task specific (domain identification, fluent response generation etc) performance in the PRALINE?

\noindent\textbf{Overall Performance on ConvQA datasets.}
Table~\ref{tab:overall_results} summarizes the results comparing PRALINE against the previous baselines. PRALINE outperforms previous baselines in all metrics on the ConvQuestions dataset. Specifically, for P@1, PRALINE performs by $0.029$ points better than CONQUER, $0.108$ points compared to CONVEX, and $0.042$ points against OAT. 
For H@5 and MRR, the margin is even more prominent, with $0.186$ and $0.100$ total points, respectively, compared against CONQUER. While for CONVEX, the margins increase to $0.310$ for H@5 and $0.198$ for MRR. For OAT, only the MRR is available, and PRALINE outperforms it by $0.138$.
For the ConvRef dataset, PRALINE performs better in all metrics compared to CONVEX, where the margin for all metrics is more than $0.100$ absolute points. Moreover, it surpasses CONQUER on ranking metrics H@5 and MRR with $0.160$ and $0.046$ points. ConvRef is an extended version of ConvQuestions with multiple question reformulations. CONQUER was sophisticatedly designed to leverage those reformulations and boost the results \cite{conquer2021kaiser}. On the other hand, PRALINE treats the reformulated questions in the same manner as it does with the original questions from the ConvQuestions benchmark. Therefore the increase for P@1 is not that significant. The effect of reformulated questions as the additional context has not been extensively studied in the scope of the work, and we leave it for future work. 

\begin{table}[!t]
\small
\begin{tabular}{l|ccc|ccc}
\toprule
\textbf{Dataset} & \multicolumn{3}{c}{\textbf{ConvQues.}} & \multicolumn{3}{c}{\textbf{ConvRef}} \\ \midrule
Model & P@1 & H@5 & MRR & P@1 & H@5 & MRR \\ \midrule
CONVEX~\cite{christmann2019look} & 0.184 & 0.219 & 0.200 & 0.225 & 0.257 & 0.241 \\ 
CONQUER \cite{conquer2021kaiser} & 0.240 & 0.343 & 0.279 & \textbf{0.353} & 0.429 & 0.387 \\
OAT \cite{marion2021structured} & 0.250 & - & 0.260 & - & - & - \\
Focal Entity \cite{lan2021modeling} & 0.248 & - & 0.248 & - & - & - \\
\midrule
\textbf{PRALINE} & \textbf{0.292} & \textbf{0.529} & \textbf{0.398} & 0.335 & \textbf{0.599} & \textbf{0.441} \\ \bottomrule
\end{tabular}
\caption{Overall results on employed datasets. The effect of incorporating conversational context in PRALINE has positively impacted empirical results, achieving better results than baselines. Best values are in bold. The total trainable parameters for PRALINE are 143M. However, OAT has 260M trainable parameters.}
\label{tab:overall_results}
\vspace{-2.5em}
\end{table}

\begin{table*}[!t]
\centering
\large
\begin{tabular}{l|cc|cc|cc|cc|cc}
\toprule
\textbf{Dataset} & \multicolumn{10}{c}{\textbf{ConvQuestions}} \\ \midrule
\textbf{Domain} & \multicolumn{2}{c}{\textbf{Movies}} & \multicolumn{2}{c}{\textbf{TV Series}} & \multicolumn{2}{c}{\textbf{Music}} & \multicolumn{2}{c}{\textbf{Books}} & \multicolumn{2}{c}{\textbf{Soccer}} \\
Models & H@5 & MRR & H@5 & MRR & H@5 & MRR & H@5 & MRR & H@5 & MRR \\ \midrule
CONVEX~\cite{christmann2019look} & 0.355 & 0.305 & 0.269 & 0.218 & 0.293 & 0.237 & 0.303 & 0.246 & 0.284 & 0.234 \\
CONQUER~\cite{conquer2021kaiser} & 0.357 & 0.316 & 0.382 & 0.325 & 0.320 & 0.263 & 0.464 & 0.417 & 0.310 & 0.268 \\ \midrule
\textbf{PRALINE} & \textbf{0.561} & \textbf{0.426} & \textbf{0.457} & \textbf{0.378} & \textbf{0.405} & \textbf{0.279} & \textbf{0.739} & \textbf{0.599} & \textbf{0.492} & \textbf{0.344} \\ \midrule
\textbf{Dataset} & \multicolumn{10}{c}{\textbf{ConvRef}} \\ \midrule
\textbf{Domain} & \multicolumn{2}{c}{\textbf{Movies}} & \multicolumn{2}{c}{\textbf{TV Series}} & \multicolumn{2}{c}{\textbf{Music}} & \multicolumn{2}{c}{\textbf{Books}} & \multicolumn{2}{c}{\textbf{Soccer}} \\
Models & H@5 & MRR & H@5 & MRR & H@5 & MRR & H@5 & MRR & H@5 & MRR \\ \midrule
CONQUER~\cite{conquer2021kaiser} & 0.436 & 0.405 & 0.442 & 0.392 & 0.398 & \textbf{0.357} & 0.554 & 0.502 & 0.360 & 0.316 \\ \midrule
\textbf{PRALINE} & \textbf{0.567} & \textbf{0.429} & \textbf{0.545} & \textbf{0.466} & \textbf{0.495} & 0.329 & \textbf{0.835} & \textbf{0.659} & \textbf{0.564} & \textbf{0.378} \\ \bottomrule
\end{tabular}
\caption{To compare the KG path ranking performance, we report fine-grained results across different domains of both benchmarks on ranking metrics. CONVEX does not report domain-specific values on ConvRef dataset, hence omitted from the respective table. PRALINE maintains an empirical edge on baselines while ranking the KG paths. Best values are in bold.}
\label{tab:convqa_results}
\vspace{-1em}
\end{table*}

\begin{table}[!t]
\begin{tabular}{lccc|ccc}
\toprule
\textbf{Dataset} & \multicolumn{3}{c}{\textbf{ConvQues.}} & \multicolumn{3}{c}{\textbf{ConvRef}} \\ \midrule
Model & P@1 & H@5 & MRR & P@1 & H@5 & MRR \\ \midrule
PRALINE & 0.292 & 0.529 & 0.398 & 0.335 & 0.599 & 0.441 \\ \midrule
w/o full conv. & 0.214 & 0.375 & 0.299 & 0.247 & 0.449 & 0.324 \\
w/o domain & 0.247 & 0.436 & 0.296 & 0.266 & 0.472 & 0.356 \\
w/o fluent resp. & 0.265 & 0.441 & 0.324 & 0.279 & 0.503 & 0.397 \\
train separately & 0.255 & 0.413 & 0.328 & 0.304 & 0.529 & 0.408 \\ \bottomrule
\end{tabular}
\caption{The effectiveness of including the entire dialog history, fluent responses, and domain information. 
The first row (from top) contains the results of PRALINE with all available contexts. 
The second row omits the full conversational history and includes only the previous turn. 
The third and forth-row selectively remove the domain information and fluent response respectively.
In the last row, we show results when we train modules independently, illustrating the advantage of joint training of PRALINE modules.}
\label{tab:ablation}
\end{table}

\noindent\textbf{Ranking Performance Across Domains.}
We further investigate the ranking performance of PRALINE across different domains for both benchmarks considering this is the main focus of our work.
Table~\ref{tab:convqa_results} illustrates detailed ranking results for H@5 and MRR as both are ranking metrics. 
As shown, for the ConvQuestions benchmark, PRALINE superiority is evident in all five domains against the baselines. However, we obtain the lowest ranking results in the \textit{Music} domain with $0.405$ for H@5 and $0.279$ for MRR. Analyzing some of the conversational examples in that domain indicated that we did not have gold positive KG paths for all the instances in the dataset. Therefore, PRALINE could not have a complete training process for all possible conversations and paths. Such issues have also impacted the baselines. 
Next, we can see that the highest-ranking results are obtained in the \textit{Books} domain, where PRALINE achieves the impressive $0.739$ for H@5, which is almost $0.300$ points higher than CONQUER. Furthermore, the highest MRR ($0.599$) is achieved in this domain. The results for this domain are heavily impacted because we managed to extract gold-standard paths for most dataset instances. Also, the number of KG relations used in positive paths is proportionately smaller than other domains, positively impacting the ranking task for all models. 
On the ConvRef benchmark, PRALINE still outperforms CONQUER on all domains. Here, PRALINE shows substantially improved results in two domains (\textit{TV series}, \textit{Books}) compared to results in ConvQuestions.
We conclude that employing contrastive learning to rank KG paths has a positive impact on the overall empirical performance of PRALINE (successfully answering \textbf{RQ)}. Furthermore, as we illustrate below in the ablation study, the conversational context (entire dialog history with fluent responses and domain information) plays a vital role in the substantially improved results.

\subsection{Ablation Study}
We perform various ablation studies on PRALINE to illustrate the effectiveness of the proposed approach and related architecture choices. Table~\ref{tab:ablation} summarizes the results of the ablation studies.

\noindent\textbf{Effect of Full Conversational History.}
Our idea here is to study the empirical advantage of incorporating entire dialog history. Hence, we created a PRALINE configuration (w/o full conv.) that only considers dialog history from the previous turn, disregarding full history. As a result, there is a drop in the performance for PRALINE (w/o full conv.), illustrated in Table~\ref{tab:ablation}.

\noindent\textbf{Effect of Domain Information.}
For the second ablation experiment, we remove the domain information (domain pointer module) from PRALINE. We can see the importance of such information in our approach. All metrics results have dropped, indicating the effect of it. In PRALINE, domain information is used to improve the conversation representation and indirectly filter out KG paths that are not relevant. Usually, such paths contain KG relations that are not used in the particular domain and add noise while ranking the correct paths.

\noindent\textbf{Effect of Fluent Response.}
To show the effectiveness of using fluent responses in conversational history, we perform an ablation experiment where we remove and replace them with standalone answers extracted from the KG (e.g. entities, literals). We can observe that the ranking performance drops significantly. In particular, we obtain a drop between $0.060-0.080$ for H@5 and MRR ranking metrics and $0.027$ for P@1. Fluent answers provide additional context in the conversational history and, therefore, support PRALINE to generate more accurate representations for the ranking task. Such context is crucial for our results. 
Hence, we conclude that conversational contexts (full dialog history with fluent responses and domain information) positively impact the KG path ranking, and it successfully answers \textbf{RQ1.1}.

\noindent\textbf{Effect of Joint Training.}
As a last ablation experiment, we study the case of whether joint training of PRALINE modules is effective. To do so, we train each module independently without sharing any sub-module parameters (e.g., encoder). Ablation results indicate lower scores for all metrics. Furthermore, we observed that all modules were overfitting much faster during this experiment than PRALINE's joint training. Intuitively, observed behavior is reasonable since the more tasks we are jointly learning; PRALINE generates better embedding representations that capture all the tasks, yielding a lower chance of overfitting.

\subsection{Detailed Architecture Analysis}
\begin{figure}[!t]
\centering
\captionsetup{type=figure}
\includegraphics[width=0.40\textwidth]{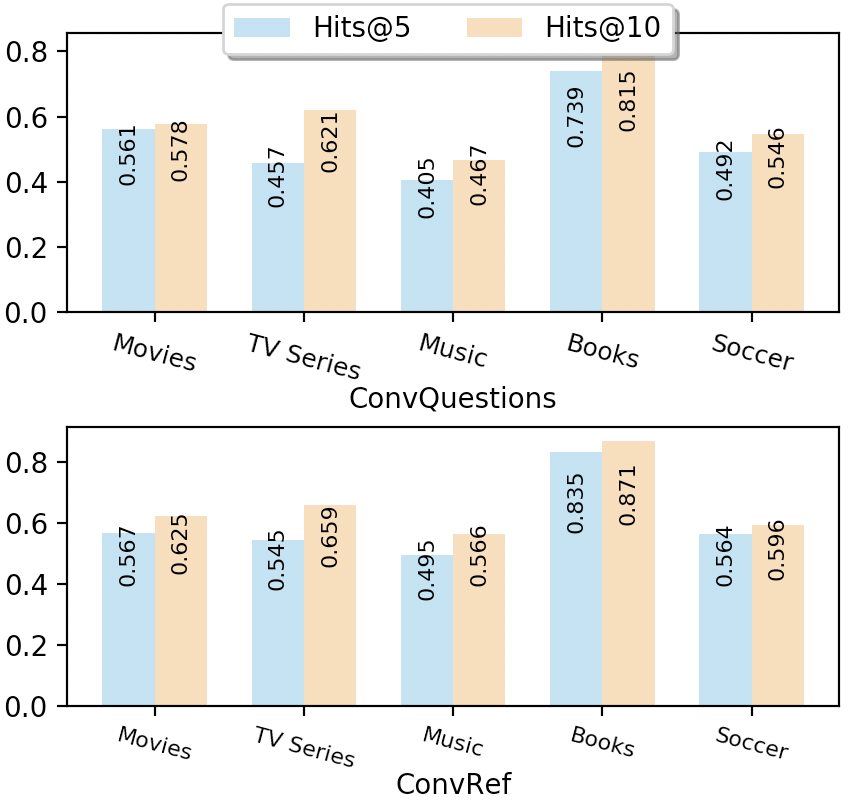}
\caption{PRALINE H@5 and H@10 ranking results.}
\label{fig:detailed_ranking_results}
\end{figure}

We calculated the task-specific performance of different modules in our framework to justify choosing various modules.
Figure~\ref{fig:detailed_ranking_results} presents PRALINE's detailed ranking results for H@5 and H@10 ranking metrics. We achieve the highest scores in the ``\textit{Books}'' domain, with H@10 of $0.871$ on the ConvRef benchmark. On the other hand, we have the lowest scores in the ``\textit{Music}'' domain. Interestingly, for most domains, results for H@5 and H@10 are relatively close, indicating that PRALINE tends to rank the correct paths in top positions, illustrating the robustness of our approach.

Table~\ref{tab:domain_results} illustrates performance of domain identification pointer on both benchmarks. For both benchmarks, the F1 score is around $0.95$. The robust results justify the use of pointer network, which also complement the ablation study in Table~\ref{tab:ablation} for PRALINE (w/o domain) configuration.

Table~\ref{tab:verbalization_results} presents the results of PRALINE's fluent response generation. We obtain a BLEU-4 score of $0.289$ on ConvQuestions and $0.327$ on ConvRef, while for METEOR, the scores are $0.626$ and $0.684$, respectively. Here, the score margins between the benchmarks are larger compared to the domain pointer task. This is due to the additional questions in ConvRef further supporting PRALINE to avoid overfitting for the response generation task. 
Here we conclude that the scores of individual modules are considered adequate and support PRALINE's performance in Tables \ref{tab:ablation} and \ref{tab:overall_results} (successfully answering \textbf{RQ1.2}).

\begin{table}[!t]
\begin{tabular}{l|ccc|ccc}
\toprule
\textbf{Dataset} & \multicolumn{3}{c}{\textbf{ConvQuestions}} & \multicolumn{3}{c}{\textbf{ConvRef}}\\ \midrule
\textbf{Task} & Pres. & Rec. & F1 & Pres. & Rec. & F1\\ \midrule
Domain Pointer & 0.951 & 0.946 & 0.947 & 0.958 & 0.959 & 0.952 \\ \bottomrule
\end{tabular}
\caption{Domain identification results.}
\label{tab:domain_results}
\end{table}

\begin{table}[!t]
\begin{tabular}{l|cc|cc}
\toprule
\textbf{Dataset} & \multicolumn{2}{c}{\textbf{ConvQuestions}} & \multicolumn{2}{c}{\textbf{ConvRef}}\\ \midrule
\textbf{Task} & BLEU-4 & METEOR & BLEU-4 & METEOR \\ \midrule
Decoder & 0.289 & 0.626 & 0.327 & 0.684 \\ \bottomrule
\end{tabular}
\caption{Fluent response generation results.}
\label{tab:verbalization_results}
\end{table}

\subsection{Error Analysis} \label{sec:error}
For the error analysis, we randomly sampled 250 incorrect predictions (with equal predictions from each domain). We detail the reasons for two types of errors observed in the analysis:

\noindent\textbf{Incorrect ranking of Paths with Semantically Similar KG Relations.}
PRALINE often wrongly-ranks paths when they contain semantically similar relations. For instance, given the question ``\textit{What kind of book is it?}'' and its entire conversation history: $q^1$) ``\textit{What is the name of the writer of The Secret Garden?}'' $v^1$) ``\textit{The name of the writer of The Secret Garden is Frances Eliza Hodgson Burnett.}'' $q^2$) ``\textit{Where does the story take place?}'' $v^2$) ``\textit{The story takes place in Yorkshire.}'' $q^3$) ``\textit{When was the book published?}'' $v^3$) ``\textit{The book was published in 1910.}''.
PRALINE is required to find the gold path that contains the KG relation "main subject (P921)" since this one points to the correct answer, which is ``\textit{adventure (Q1436734)}''. However, PRALINE here ranks higher paths that contain the KG relation ``\textit{genre (P136)}''. For the example mentioned above, there are three KG paths with relation ``\textit{genre}'', and all of them are ranked in the top three positions. As we can see, the relations ``\textit{main subject}'' and ``\textit{genre}'' are semantically similar and, therefore, hard to distinguish which one to rank higher. For the particular example, the relation ``\textit{genre}'' is ranked higher since it is used more across the gold KG paths in training data. In this work, we focused on context derived from the conversation and have not considered widely available KG context such as entity/relation aliases, types, etc. 

\noindent\textbf{Absence of Gold KG Paths.}
Several examples (over $25\%$) with missing gold paths in training datasets significantly affect the learning process.
For the test sets, there were $19\%$ of conversational turns without gold KG paths. These examples are directly counted as wrong instances and negatively affect our results. With a more sophisticated annotation process for gold KG paths, PRALINE results would have been improved.

\enlargethispage{2\baselineskip}
\section{Conclusions and Future Work}\label{sec:conclusion}
Our central research question was to study the impact of contrastive representation learning for conversational question answering over knowledge graphs. To accomplish this, we formulate the task as a KG path ranking problem, and we leverage conversational contexts such as full dialog history with fluent responses and domain information. 
The approach implemented in the PRALINE framework and its associated empirical advantage over baselines provides conclusive evidence of the effectiveness of the contrastive learning for the task. We conclude that a joint embedding of conversation and KG-paths in a homogeneous space positively impacts the overall ranking metrics. Furthermore, our systematic ablation studies illustrate each conversation context's impact (entire conversation history, fluent responses, and domain information) on PRALINE's performance. Based on our findings, extensive evaluations, and gained insights in this paper, we point readers to the following future research directions: 1) The error analysis shows limitations in finding the correct KG relation. Relation extraction and linking it to KGs for conversations is a key open research question. It would be a logical next step to incorporate KG context for relation extraction similar to RECON~\cite{bastos2021recon}. 2) The hit@5 results are relatively high compared to precision@1 for PRALINE and baselines, with an expansive room for empirical improvement by proposing new approaches. We believe our findings will pave the way for more research on this relatively unexplored IR task in the ConvQA domain.

\newpage



\enlargethispage{\baselineskip}

\end{document}